\documentclass[11pt,a4paper]{article}
\usepackage[hyperref]{emnlp-ijcnlp-2019}

\usepackage{times}
\usepackage{soul}
\usepackage{CJKutf8}
\usepackage{url}
\usepackage{amsmath}
\usepackage[utf8]{inputenc}
\usepackage{graphicx}
\usepackage{amsmath}
\usepackage{booktabs}
\usepackage{xcolor}
\usepackage{algorithm}
\usepackage{algorithmic}
\usepackage{amssymb}
\usepackage{makecell}
\usepackage{xspace}
\usepackage{array}
\usepackage[normalem]{ulem}
\usepackage{footmisc}
\usepackage{colortbl}
\usepackage{soul}
\usepackage{epstopdf}
\usepackage{setspace}
\usepackage{pifont}
\usepackage{mathrsfs}
\usepackage{subfig}

\newcommand{\ignore}[1]{}

\newcommand{\dubbelop}{$^{\blacktriangle}$}

\newcommand{\dubbelneer}{$^{\blacktriangledown}$}

\newcommand\icst{$^\spadesuit$}
\newcommand\tencent{$^\diamondsuit$}
\newcommand\data{$^\clubsuit$}

\makeatletter 
  \newcommand\figcaption{\def\@captype{figure}\caption} 
  \newcommand\tabcaption{\def\@captype{table}\caption} 
\makeatother
\title{How to Write Summaries with Patterns? \\ Learning towards Abstractive Summarization through Prototype Editing}
\aclfinalcopy
\author{
Shen Gao \icst\thanks{\;\;Equal contribution. Ordering is decided by a coin flip.}, 
Xiuying Chen \icst\data\footnotemark[1], 
Piji Li \tencent, 
Zhangming Chan \icst\data, 
Dongyan Zhao \icst\data,  
Rui Yan \icst\data\thanks{\;\;Corresponding author.} \\
\icst Wangxuan Institute of Computer Technology, Peking University, Beijing, China, \\
\tencent Tencent AI Lab, Shenzhen, China\\ 
\data Center for Data Science, Peking University, Beijing, China\\
{\tt \{shengao, xy-chen, zhangming.chan, zhaody, ruiyan\}@pku.edu.cn} \\
{\tt pijili@tencent.com}
}

\date{}
\begin{document}

\maketitle

\begin{abstract}
Under special circumstances, summaries should conform to a particular style with patterns, such as court judgments and abstracts in academic papers.
To this end, the prototype document-summary pairs can be utilized to generate better summaries.
There are two main challenges in this task:
(1) the model needs to incorporate learned patterns from the prototype, but (2) should avoid copying contents other than the patternized words---such as irrelevant facts---into the generated summaries.
To tackle these challenges, we design a model named \textit{Prototype Editing based Summary Generator} (PESG).
PESG first \textbf{learns summary patterns} and prototype facts by analyzing the correlation between a prototype document and its summary.
Prototype facts are then utilized to help \textbf{extract facts} from the input document.
Next, an editing generator generates new summary based on the \textbf{summary pattern} or \textbf{extracted facts}. 
Finally, to address the second challenge, a fact checker is used to estimate mutual information between the input document and generated summary, providing an additional signal for the generator.
Extensive experiments conducted on a large-scale real-world text summarization dataset\footnote{\url{https://github.com/gsh199449/proto-summ}} show that PESG achieves the state-of-the-art performance in terms of both automatic metrics and human evaluations. 
\end{abstract}

\section{Introduction}

Abstractive summarization can be regarded as a sequence mapping task that maps the source text to the target summary \cite{rush2015neural,li2017deep,Cao2018Retrieve,Gao2019Abstractive}.
It has drawn significant attention since the introduction of deep neural networks to natural language processing.
Under special circumstances, the generated summaries are required to conform to a specific pattern, such as court judgments, diagnosis certificates, abstracts in academic papers, etc.
Take the court judgments for example, there is always a statement of the crime committed by the accused, followed by the motives and the results of the judgment.
An example case is shown in Table~\ref{tab:intro-case}, where the summary shares the same writing style and has words in common with the prototype summary (retrieved from the training dataset).

\begin{table}[t]
    \centering
    \scriptsize
    \begin{tabular}{l|l}
        \toprule
        \rotatebox{270}{prototype summary} & \multicolumn{1}{p{6.6cm}}{        \textcolor{red}{The court held that the} \textcolor{blue}{defendant Wang had } \textcolor{red}{stolen the property of others for the purpose of illegal possession. The amount was large, and his behavior constituted the crime of theft.} \textcolor{red}{The accusation of the public prosecution agency was established.} \textcolor{blue}{The defendant Wang has a criminal record and will be considered when sentencing. Since the defendant Wang did not succeed because of reasons other than his will, he could be punished lightly. After the defendant confessed his crimes to the case, he was} \textcolor{red}{given a lighter punishment according to law}.
        } \\ \hline 
        \rotatebox{270}{summary} & \multicolumn{1}{p{6.6cm}}{
            \textcolor{red}{The court held that} \textcolor{blue}{ the accused Zhang and Fan} \textcolor{red}{stole property and the amount was large. Their actions constituted the crime of theft.} \textcolor{red}{The accusation of the public prosecution agency was established} \textcolor{blue}{and supported. This crime was committed within two years after the release of the defendants Zhang and Fan. Thus they are recidivists and this situation will be considered when sentencing. The fact that defendants Zhang and Fan surrendered themselves and pleaded guilty in court} \textcolor{red}{gives a lighter punishment according to law}.
        }   \\
        \bottomrule
    \end{tabular}
    \caption{An example of patternized summary generation. The text in red denotes patternized words shared in different summaries, and text in blue denotes specific facts. 
    }
    \label{tab:intro-case}
\end{table}

Existing prototype based generation models such as~\cite{Wu2018ResponseGB} are all applied on short text, thus, cannot handle long documents summarization task.
Another series of works focus on template-based methods such as \cite{oya2014template}.
However, template-based methods are too rigid for our patternized summary generation task.
Hence, in this paper, we propose a summarization framework named \emph{Prototype Editing based Summary Generator} (PESG) that incorporates prototype document-summary pairs to improve summarization performance when generating summaries with pattern.
First, we calculate the cross dependency between the prototype document-summary pair to obtain a summary pattern and prototype facts (explained in \S~\ref{sec:proto-reader}).
Then, we extract facts from the input document with the help of the prototype facts (explained in \S~\ref{memory}).
Next, a recurrent neural network (RNN) based decoder is used to generate a new summary, incorporating both the summary pattern and extracted facts (explained in \S~\ref{sec:edit-gen}). 
Finally, a fact checker is designed to provide mutual information between the generated summary and the input document to prevent the generator from copying irrelevant facts from the prototype (explained in \S~\ref{sec:factchecker}).
To evaluate PESG, we collect a large-scale court judgment dataset, where each judgment is a summary of the case description with a patternized style.
Extensive experiments conducted on this dataset show that PESG outperforms the state-of-the-art summarization baselines in terms of ROUGE metrics and human evaluations by a large margin.

Our contributions can be summarized as follows: 

$\bullet$ We propose to use prototype information to help generate better summaries with patterns.

$\bullet$ Specifically, we propose to generate the summary incorporating the prototype summary pattern and extracted facts from input document.

$\bullet$ We provide mutual information signal for the generator to prevent copying irrelevant facts from the prototype.

$\bullet$ We release a large-scale prototype based summarization dataset that is beneficial for the community.
\section{Related Work}
We detail related work on text summarization and prototype editing.

Text summarization can be classified into extractive and abstractive methods. 
Extractive methods~\cite{Narayan2018RankingSF,chen2018iterative} directly select salient sentences from an article to compose a summary.
One shortcoming of these models is that they tend to suffer from redundancy.
Recently, with the emergence of neural network models for text generation, a vast majority of the literature on summarization~\cite{Ma2018AutoencoderAA,Zhou2018SequentialCN,Gao2019Abstractive,chen2019Timeline} is dedicated to abstractive summarization, which aims to generate new content that concisely paraphrases a document from scratch.

Another line of research focuses on prototype editing.
\cite{Guu2018GeneratingSB} proposed the first prototype editing model, which samples a prototype sentence from training data and then edits it into a new sentence.
Following this work,
\cite{Wu2018ResponseGB} proposed a new paradigm for response generation, which first retrieves a prototype response from a pre-defined index and then edits the prototype response.
\cite{Cao2018Retrieve} applied this method on summarization, where they employed existing summaries as soft templates to generate new summary without modeling the dependency between the prototype document, summary and input document.
Different from these soft attention methods, \cite{cai2018skeleton} proposed a hard-editing skeleton-based model to promote the coherence of generated stories. 
Template-based summarization is also a hard-editing method \cite{oya2014template}, where a multi-sentence fusion algorithm is extended in order to generate summary templates.

Different from all above works, our model focuses on patternized summary generation, which is more challenging than traditional news summarization and short sentence prototype editing. %
\section{Problem Formulation}

For an input document $X = \{x_1, x_2, \dots, x_{T_m}\}$, we assume there is a ground truth summary $Y = \{y_1, y_2, \dots, y_{T_n}\}$.
In our prototype summarization task, a retrieved prototype document $\hat{X} = \{\hat{x}_1, \hat{x}_2, \dots, \hat{x}_{T_m}\}$ with a corresponding prototype summary $\hat{Y} = \{\hat{y}_1, \hat{y}_2, \dots, \hat{y}_{T_n}\}$ is also attached according to their similarities with $X$.

For a given document $X$, our model extracts salient facts from $X$ guided by a prototype document $\hat{X}$, and then generates the summary $Y^{'}$ by referring to the prototype summary $\hat{Y}$.
The goal is to generate a summary $Y^{'}$ that not only follows a patternized style (as defined by prototype summary $\hat{Y}$) but also is consistent with the facts in document $X$.

\section{Model}

\subsection{Overview}

\begin{figure}
    \centering
    \includegraphics[scale=0.35]{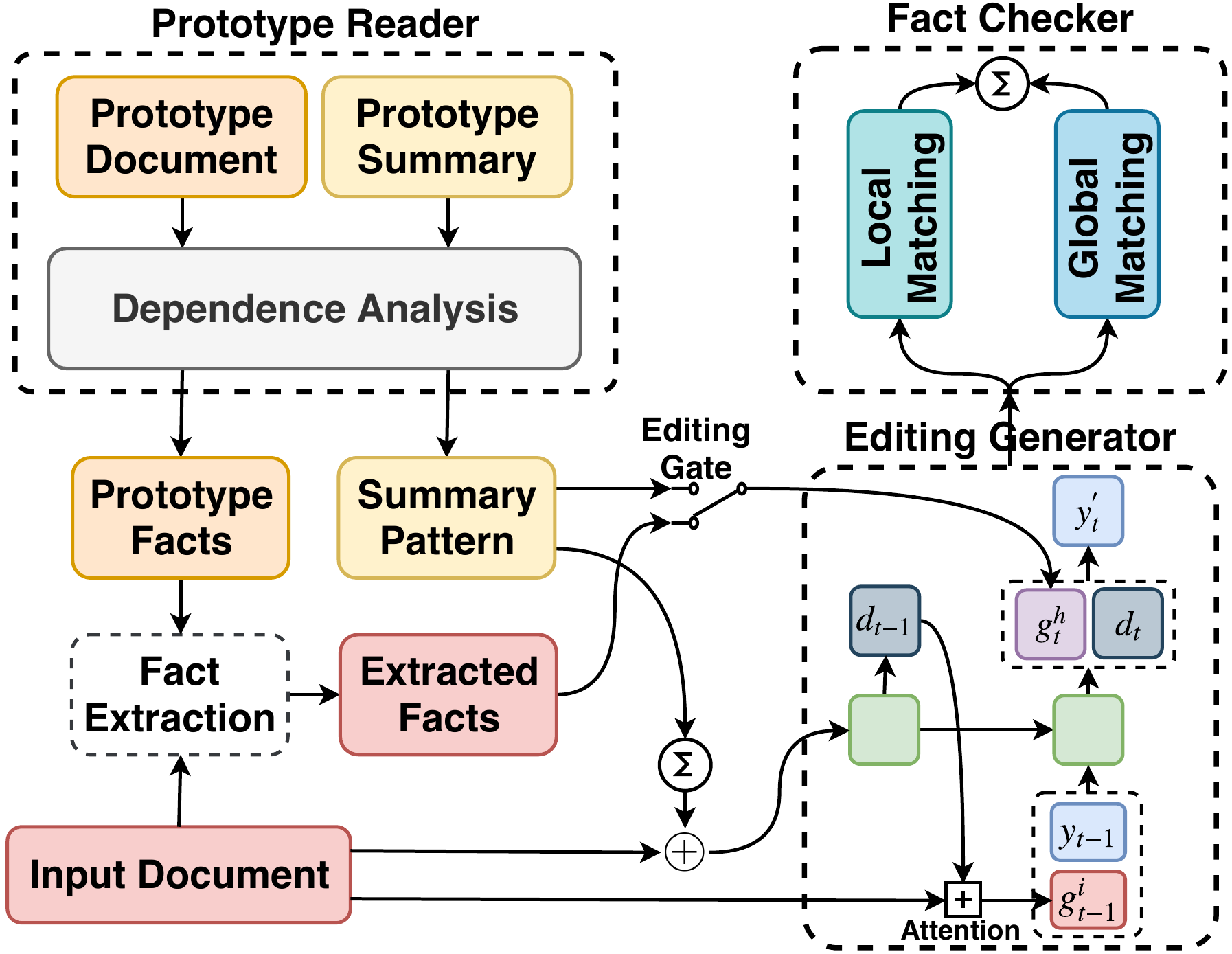}
    \caption{
    Overview of PESG. We divide our model into four parts: (1) \textit{Prototype Reader}; (2) \textit{Fact Extraction}; (3) \textit{Editing Generator}; (4) \textit{Fact Checker}.
    }
    \label{fig:overview}
\end{figure}

In this section, we propose our \emph{prototype editing based summary generator}, which can be split into two main parts, as shown in Figure~\ref{fig:overview}:

$\bullet$ \textit{Summary Generator.}
(1) \textbf{Prototype Reader} analyzes the dependency between $\hat{X}$ and $\hat{Y}$ to determine the summary pattern and prototype facts.
(2) \textbf{Fact Extraction} module extracts facts from the input document under the guidance of the prototype facts.
(3) \textbf{Editing Generator} module generates the summary $Y^{'}$ of document $X$ by incorporating summary pattern and facts.

$\bullet$ \textit{Fact Checker} estimates the mutual information between the generated summary $Y^{'}$ and input document $X$. This information provides an additional signal for the generation process, preventing irrelevant facts from being copied from the prototype document.

\subsection{Prototype Reader}
\label{sec:proto-reader}

To begin with, we use an embedding matrix $e$ to map a one-hot representation of each word in $X$, $\hat{X}$, $\hat{Y}$ into a high-dimensional vector space.
We then employ a bi-directional recurrent neural network (Bi-RNN) to model the temporal interactions between words:
\begin{align}
    h^x_t &= \text{Bi-RNN}_x(e(x_t), h^x_{t-1}), \\
    \hat{h}^x_t &= \text{Bi-RNN}_x(e(\hat{x}_t), \hat{h}^x_{t-1}), \\
    \hat{h}^y_t &= \text{Bi-RNN}_y(e(\hat{y}_t), \hat{h}^y_{t-1}),
\end{align}
\noindent where $h^x_t$, $\hat{h}^x_t$ and $\hat{h}^y_t$ denote the hidden state of $t$-th step in Bi-RNN for $X$, $\hat{X}$ and $\hat{Y}$, respectively. 
Following~\cite{Tao2018GetTP,gao2019product,Hu2019GSN}, we choose long short-term memory (LSTM) as the cell for Bi-RNN. 

On one hand, the sections in the prototype summary that are not highly related to the prototype document are the universal patternized words and should be emphasized when generating the new summary.
On the other hand, the sections in the prototype document that are highly related to the prototype summary are useful facts that can guide the process of extracting facts from input document.
Hence, we employ a bi-directional attention mechanism between a prototype document and summary to analyze the cross-dependency, that is, from document to summary and from summary to document.
Both of these attentions are derived from a shared similarity matrix, $S \in \mathbb{R}^{T_m \times T_n}$, calculated by the hidden states of prototype document $\hat{X}$ and prototype summary $\hat{Y}$.
$S_{ij}$ indicates the similarity between the $i$-th document word $\hat{x}_i$ and $j$-th summary word $\hat{y}_j$ and is computed as:
\begin{equation}
\begin{aligned}
    S_{ij} &= \alpha(\hat{h}^x_i, \hat{h}^y_j), \\
    \alpha(x, y) &= w^\intercal [x \oplus y \oplus (x \otimes y)] ,
\end{aligned}
\label{eq:alpha}
\end{equation}
where $\alpha$ is a trainable scalar function that calculates the similarity between two input vectors. $\oplus$ denotes a concatenation operation and $\otimes$ is an element-wise multiplication.

We use $a_t^s=\text{mean}(S_{:t}) \in \mathbb{R}$ to represent the attention weight on the $t$-th prototype summary word by document words, which will learn to assign high weights to highly related universal patternized words when generating a summary.
From $a_t^s$, we obtain the weighted sum of the hidden states of prototype summary as ``\textbf{summary pattern}'' $l = \{l_1, \dots, l_{T_n}\}$, where $l_i$ is:
\begin{equation}
l_i = a^s_i \hat{h}^y_i.
\end{equation}

Similarly, $a_t^d=\text{mean}(S_{t:}) \in \mathbb{R}$ assigns high weights to the words in a prototype document that are relevant to the prototype summary.
A convolutional layer is then applied to extract ``\textbf{prototype facts}'' $\hat{r}_{t}$ from the prototype document:
\begin{equation}
\hat{r}_{t} = \text{CNN}(a^d_t \hat{h}^x_t). \label{proto-facts}
\end{equation}
We sum the prototype facts to obtain the overall representation of these facts:
\begin{equation}
    q = \textstyle \sum_{t}^{T_m} \hat{r}_{t}. \label{proto-doc-repr}
\end{equation}

\subsection{Fact Extraction}
\label{memory}

In this section, we discuss how to extract useful facts from an input document with the help of prototype facts.

We first extract the facts from an input document by calculating their relevance to prototype facts.
The similarity matrix $E$ is then calculated between the weighted prototype document $a^d_i \hat{h}^x_i$ and input document representation $h^x_j$:
\begin{equation}
    E_{ij} = \alpha(a^d_i \hat{h}^x_i, h^x_j),
\end{equation}
where $\alpha$ is the similarity function introduced in Equation~\ref{eq:alpha}.
Then, we sum up $E_{ij}$ along the length of the prototype document to obtain the weight $E_{j} = \sum_{ti}^{T_m} E_{tj}$ for $j$-th word in the document.
Next, similar to Equation~\ref{proto-facts}, a convolutional layer is applied on the weighted hidden states $E_{t} h^x_t$ to obtain the fact representation $r_{t}$ from the input document:
\begin{equation}
    r_{t} = \text{CNN}(E_{t} h^x_t). \label{doc-facts}
\end{equation}

\begin{figure}
    \centering
    \includegraphics[scale=0.35]{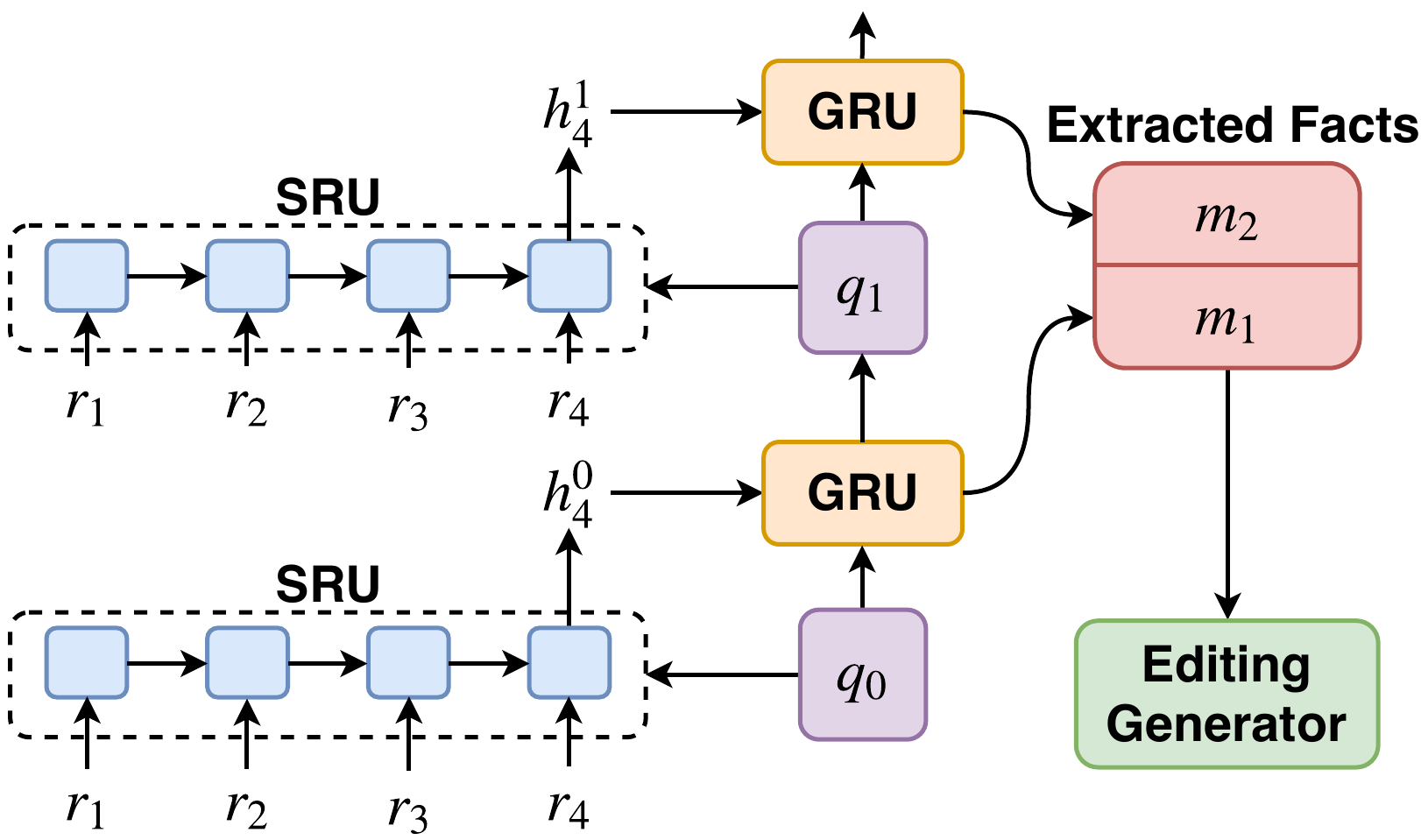}
    \caption{Framework of fact extraction module.}
    \label{fig:memory-framework}
\end{figure}

Inspired by the polishing strategy in extractive summarization~\cite{chen2018iterative}, we propose to use the prototype facts to polish the extracted facts $r_{t}$ and obtain the final fact representation $m_{.}$, as shown in Figure~\ref{fig:memory-framework}.
Generally, the polishing process consists of two hierarchical recurrent layers.
The first recurrent layer is made up of Selective Recurrent Units (SRUs), which take facts $r_{\cdot}$ and polished fact $q_k$ as input, outputting the hidden state $h^k_{T_m}$.
The second recurrent layer consists of regular Gated Recurrent Units (GRUs), which are used to update the polished fact from $q_k$ to $q_{k+1}$ using $h^k_{T_m}$.

SRU is a modified version of the original GRU introduced in ~\cite{chen2018iterative}, details of which can be found in Appendix~\ref{sec:fgru}.
Its difference from GRU lies in that the update gate in SRU is decided by both the polished fact $q_{k}$ and original fact $r_{t}$ together.
The $t$-th hidden state of SRU is calculated as:
\begin{align}
    h^{k}_t &= \text{SRU}(r_t, q_{k}).
\end{align}
We take $h^{k}_{T_m}$ as the overall representation of all input facts $r_\cdot$.
In this way, SRU can decide to which degree each unit should be updated based on its relationship with the polished fact $q_{k}$.

Next, $h^{k}_{T_m}$ is used to update the polished fact $q_k$ using the second recurrent layer, consisting of GRUs:
\begin{equation}
    m_{k+1}, q_{k+1} = \text{GRU}(h^{k}_{T_m}, q_{k}),
\end{equation}
where $q_{k}$ is the cell state, $h^{k}_{T_m}$ is the input and $m_{k+1}$ is the output hidden state.
$q_0$ is initialized using $q$ in Equation~\ref{proto-doc-repr}.
This iterative process is conducted $K$ times, and each output $m_k$ is stored as \textbf{extracted facts} $M = \{m_1, m_2, \dots, m_K\}$.
In this way, $M$ stores facts with different polished levels.

\subsection{Editing Generator} \label{sec:edit-gen}

The editing generator aims to generate a summary based on the input document, prototype summary and extracted facts. 
As with the settings of prototype reader, we use LSTM as the RNN cell.
We first apply a linear transformation on the summation of the summary pattern $l^{'} = \textstyle \sum_i^{T_n} l_i$ and input document representations $h^x_{T_m}$, and then employ this vector as the initial state $d_0$ of the RNN generator as shown in Equation~\ref{init-state}.
The procedure of $t$-th generation is shown in Equation~\ref{dec-step}:
\begin{align}
    d_0 &= W_e [ h^x_{T_m} \oplus l^{'} ] + b_e , \label{init-state} \\
    d_t &= \text{LSTM} (d_{t-1}, [g^i_{t-1} \oplus e(y_{t-1})]) , \label{dec-step}
\end{align}
where $W_e, b_e$ are trainable parameters, $d_t$ is the hidden state of the $t$-th generating step, and $g_{t-1}^{i}$ is the context vector produced by the standard attention mechanism~\cite{Bahdanau2014NeuralMT}.

To take advantage of the extracted facts $M$ and prototype summary $l$, we incorporate them both into summary generation using a dynamic attention.
More specifically, we utilize a matching function $f$ to model the relationship between the current decoding state $d_t$ and each $v_i$ ($v_i$ can be a extracted fact $m_i$ or summary pattern $l_i$):
\begin{align}
    \delta_{it} &= \frac{\exp(f(v_i, d_t))}{\sum_{j}^{K} \exp(f(v_j, d_t))} , \\
    g^{*}_t &= \textstyle \sum_{i}^{K} \delta_{it} v_i, \label{mem-context}
\end{align}
where $g^{*}_t$ can be $g^m_t$ or $g^s_t$ for attending to extracted facts or a summary pattern, respectively.
We use a simple but efficient bi-linear layer as the matching function $f = m_i W_f d_t$.
As for combining $g^m_t$ and $g^s_t$, we propose to use an ``editing gate'' $\gamma$, which is determined by the decoder state $d_t$, to decide the importance of the summary pattern and extracted facts at each decoding step.
\begin{equation}
    \gamma = \sigma \left( W_g d_t + b_g \right) , \label{editing-gate}
\end{equation}
where $\sigma$ denotes the sigmoid function.
Using the editing gate, we obtain $g^h_t$ which dynamically combines information from the extracted facts and summary pattern with the editing gate $\gamma$, as:
\begin{equation}
    g^h_t = \left[ \gamma g^m_t \oplus \left( 1 - \gamma \right) g^s_t \right] . \label{context}
\end{equation}
Finally, the context vector $g^h_t$ is concatenated with the decoder state $d_t$ and fed into a linear layer to obtain the generated word distribution $P_{v}$:
\begin{align}
    d^o_t &= W_o [d_t \oplus g^h_t] + b_o  \label{dec-linear},\\
    P_{v} &= \text{softmax} \left(W_v d^o_t + b_v \right).
\end{align}
The loss is the negative log likelihood of the target word $y_t$:
\begin{equation}\label{loss-s}
    \mathcal{L}_s = - \textstyle \sum^{T_n}_{t=1} \log P_{v}(y_t).
\end{equation}

In order to handle the out-of-vocabulary (OOV) problem, we equip our decoder with a pointer network~\cite{Gu2016IncorporatingCM,vinyals2015pointer,See2017GetTT}.
This process is the same as the model described in~\cite{See2017GetTT}, thus, is omit here due to limited space.

What's more, previous work~\cite{Holtzman2018LearningTW} has found that using a cross entropy loss alone is not enough for generating coherent text.
Similarly, in our task, using $\mathcal{L}_s$ alone is not enough to distinguish a good summary with accurate facts from a bad summary with detailed facts from the prototype document (see \S~\ref{sec:ablation}).
Thus, we propose a fact checker to determine whether the generated summary is highly related to the input document.

\subsection{Fact Checker}
\label{sec:factchecker}
\begin{figure}
    \centering
    \includegraphics[scale=0.30]{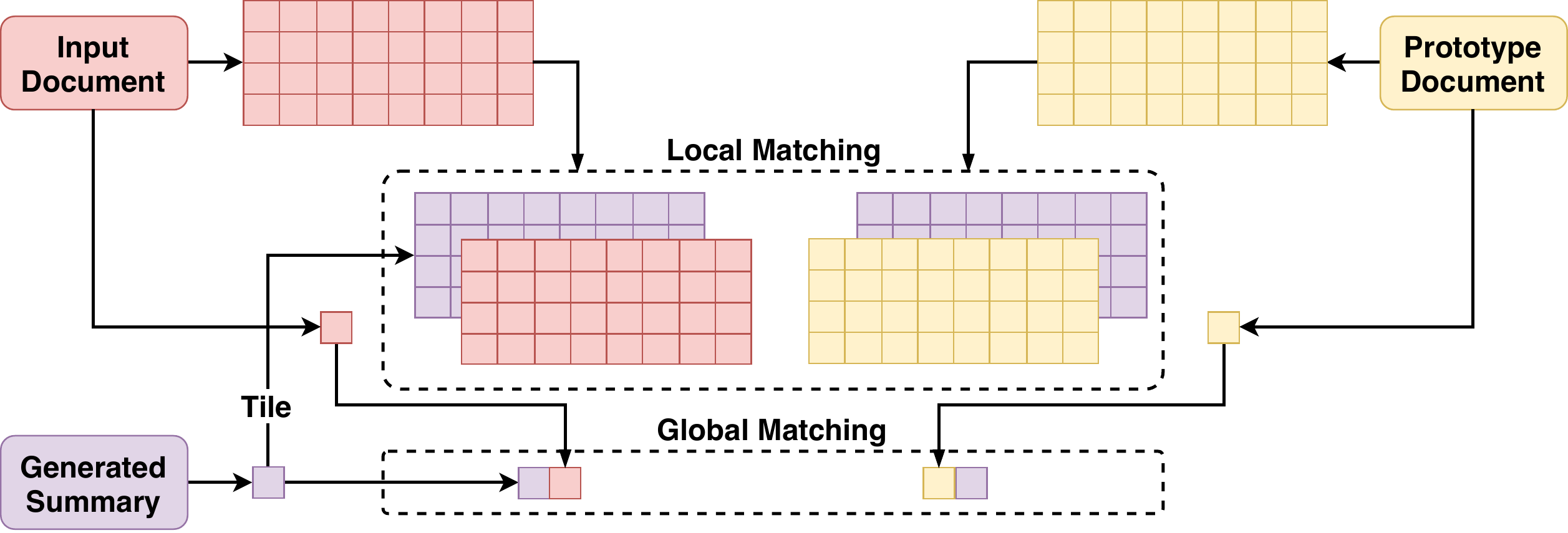}
    \caption{Framework of fact checker module.}
    \label{fig:mutual}
\end{figure}

To generate accurate summaries that are consistent with the detailed facts from the input document rather than facts from the prototype document, we add a fact checker to provide additional training signals for the generator.
Following \cite{Hjelm2018LearningDR}, we employ the neural mutual information estimator to estimate the mutual information between the generated summary $Y^{'}$ and its corresponding document $X$, as well as the prototype document $\hat{X}$.
Generally, mutual information is estimated from a local and global level, and we expect the matching degree to be higher between the generated summary and input document than the prototype document.
An overview of the fact checker is shown in Figure~\ref{fig:mutual}. 

To begin, we use a local matching network to calculate the matching degree, for local features, between the generated summary and the input, as well as prototype document.
Remember that, in \S~\ref{memory}, we obtain the fact representation of an input document $r_{i}$ and prototype facts $\hat{r}_{i}$. 
Combining these with the final hidden state $d_{T_n}$ of the generator RNN (in Equation~\ref{dec-step}), yields the local features of input extracted facts and the prototype facts:
\begin{align}
    C^r = \{d_{T_n} \oplus r_1, \dots, d_{T_n} \oplus r_{T_m}\}, \\
    C^f = \{d_{T_n} \oplus \hat{r}_1, \dots, d_{T_n} \oplus \hat{r}_{T_m}\}.
\end{align}
A $1 \times 1$ convolutional layer and a fully-connected layer are applied to score these two features:
\begin{equation}
    \tau^r_l = \text{CNN}_l(C^r), \quad \tau^f_l = \text{CNN}_l(C^f) ,
\end{equation}
where $\tau^r_l \in \mathbb{R}$, $\tau^f_l \in \mathbb{R}$ represent the local matching degree between the generated summary and input document and prototype document, respectively.
We want the generated summary to be more similar to the input document than the prototype document. 
Thus, the optimization objective of the local matching network is to minimize $\mathcal{L}_l$:
\begin{equation}
    \mathcal{L}_l = - \left( \log(\tau^r_l) + \log(1-\tau^f_l) \right).
\end{equation}

We also have a global matching network to measure the matching degree, for global features, between the generated summary and the input document, as well as prototype document.
To do so, we concatenate the representation of the generated summary with the final hidden state of the input document $h^x_{T_m}$ and final state of the prototype document $\hat{h}^x_{T_m}$, respectively, and apply a linear layer to these:
\begin{align}
    \tau^r_g &= \text{relu}(W_m [d_{T_n} \oplus h^x_{T_m}] + b_m),\\
    \tau^f_g &= \text{relu}(W_m [d_{T_n} \oplus \hat{h}^x_{T_m}] + b_m),
\end{align}
where $W_m, b_m$ are trainable parameters and $\tau^r_g \in \mathbb{R}$ and $\tau^f_g \in \mathbb{R}$ represent the matching degree between the generated summary and the input document, and prototype document, respectively.
The objective of this global matching network, similar to the local matching network, is to minimize:
\begin{equation}
    \mathcal{L}_g = - \left( \log(\tau^r_g) + \log(1-\tau^f_g) \right) .
\end{equation}
Finally, we combine the local and global loss functions to obtain the final loss $\mathcal{L}$, which we use $\mathcal{L}$ to calculate the gradients for all parameters:
\begin{equation}
    \mathcal{L} = \epsilon \mathcal{L}_g + \eta \mathcal{L}_l + \mathcal{L}_s , \label{total-loss}
\end{equation}
where $\epsilon, \eta$ are both hyper parameters.
To optimize the trainable parameters, we employ the gradient descent method Adagrad~\cite{Duchi2010AdaptiveSM} to update all parameters.
\section{Experimental Setup}

\subsection{Dataset}

We collect a large-scale prototype based summarization dataset\footnote{\url{https://github.com/gsh199449/proto-summ}}, which contains 2,003,390 court judgment documents.
In this dataset, we use a case description as an input document and the court judgment as the summary. 
The average lengths of the input documents and summaries are 595.15 words and 273.57 words respectively.
The percentage of words common to a prototype summary and the reference summary is 80.66\%, which confirms the feasibility and necessity of prototype summarization.
Following other summarization datasets~\cite{Grusky2018NewsroomAD,Kim2019Abstractive,Narayan2018DontGM}, we also count the novel n-grams in a summary compared with the n-grams in the original document, and the percentage of novel n-grams are 51.21\%, 84.59\%, 91.48\%, 94.83\% for novel 1-grams to 4-grams respectively. 
The coverage, compression and density~\cite{Grusky2018NewsroomAD} are commonly used as metrics to evaluate the abstractness of a summary.
For the summaries in our dataset, the coverage percentage is 48.78\%, compression is 2.28 and density is 1.31.
We anonymize entity tokens into special tags, such as using ``PERS'' to replace a person's name.

\begin{figure*}[ht]
  \centering
  \subfloat{ 
    \includegraphics[clip, width=1.8\columnwidth]{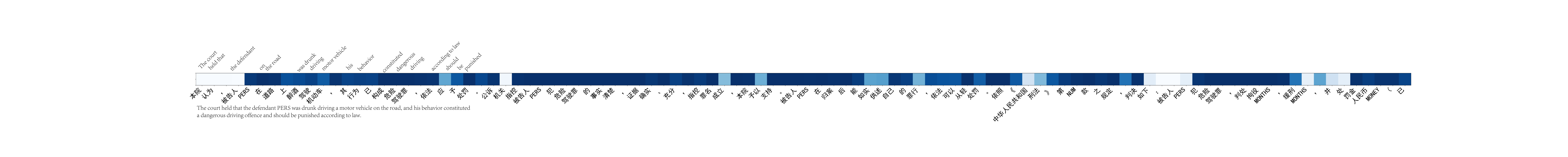}}

    \caption{Visualizations of editing gate.}
  \label{fig:edit-gate-vis}
\end{figure*}

\subsection{Comparisons}

\begin{table}[t]
    \centering
    \scriptsize
    \begin{tabular}{@{}l@{~}l}
        \toprule
        Acronym & Gloss \\
        \midrule
        PESG-FC &  \multicolumn{1}{p{3cm}}{PESG w/o Fact Checker}\\
        PESG-PR &  \multicolumn{1}{p{3cm}}{PESG w/o Prototype Reader}\\
        PESG-SS &  \multicolumn{1}{p{3cm}}{PESG w/o Summary Pattern}\\
        PESG-FG &  \multicolumn{1}{p{3cm}}{PESG w/o FGRU}\\
        \bottomrule
    \end{tabular}
    \caption{Ablation models for comparison.}
    \label{tab:ablations}
\end{table}

In order to prove the effectiveness of each module of PESG, we conduct several ablation studies, shown in Table~\ref{tab:ablations}.
We also compare our model with the following baselines:
(1) \texttt{Lead-3} is a commonly used summarization baseline~\cite{Nallapati2017SummaRuNNerAR,See2017GetTT}, which selects the first three sentences of document as the summary.
(2) \texttt{S2S} is a sequence-to-sequence framework with a pointer network, proposed by \cite{See2017GetTT}. 
(3) \texttt{Proto} is a context-aware prototype editing dialog response generation model proposed by \cite{Wu2018ResponseGB}.
(4) \texttt{Re\textsuperscript{3}Sum}, proposed by \cite{Cao2018Retrieve}, uses an IR platform to retrieve proper summaries and extends the seq2seq framework to jointly conduct template-aware summary generation.
(5) \texttt{Uni-model} was proposed by \cite{hsu2018unified}, and is the current state-of-the-art abstractive summarization approach on the CNN/DailyMail dataset.
(6) We also directly concatenate the prototype summary with the original document as input for S2S and Uni-model, named as \texttt{Concat-S2S} and \texttt{Concat-Uni}, respectively.

\subsection{Evaluation Metrics}

For the court judgment dataset, we evaluate standard ROUGE-1, ROUGE-2 and ROUGE-L~\cite{lin2004rouge} on full-length F1 following previous works~\cite{Nallapati2017SummaRuNNerAR,See2017GetTT,paulus2018a}, where ROUGE-1 (R1), ROUGE-2 (R2), and ROUGE-L (RL) refer to the matches of unigram, bigrams, and the longest common subsequence respectively.

\cite{E17-2007} notes that only using the ROUGE metric to evaluate summarization quality can be misleading. 
Therefore, we also evaluate our model by human evaluation.
Three highly educated participants are asked to score 100 randomly sampled summaries generated by three models: \texttt{Uni-model}, \texttt{Re\textsuperscript{3}Sum} and PESG. 
The statistical significance of observed differences between the performance of two runs is tested using a two-tailed paired t-test and is denoted using \dubbelop (or \dubbelneer) for strong (or weak) significance for $\alpha = 0.01$. 

\subsection{Implementation Details}

We implement our experiments in TensorFlow~\cite{abadi2016tensorflow} on an NVIDIA GTX 1080 Ti GPU. 
The word embedding dimension is 256 and the number of hidden units is 256.
The batch size is set to 64.
We padded or cut input document to contain exactly 250 words, and the decoding length is set to 100.
$\epsilon$ and $\eta$ from the Equation~\ref{total-loss} are both set to 1.0.
We initialize all of the parameters randomly using a Gaussian distribution.
We use Adagrad optimizer~\cite{Duchi2010AdaptiveSM} as our optimizing algorithm and employ beam search with size 5 to generate more fluency summary sentence.
We also apply gradient clipping~\cite{Pascanu2013OnTD} with range $[-5, 5]$ during training. 
We use dropout~\cite{Srivastava2014DropoutAS} as regularization with keep probability $p = 0.7$.

\section{Experimental Result}

\subsection{Overall Performance}

\newcommand{\cbkgrnd}{\cellcolor{blue!15}}
\newcommand{\sbbkgrnd}{\cellcolor{gray!65}}
\newcommand{\phantomtriangle}{\phantom{\dubbelop}}

\begin{table}[t]
\centering
\small
\begin{tabular}{@{}lccc@{}}
\toprule
& R1 & R2 & RL\\
\midrule
Lead-3     & 32.9 & 13.3 & 30.0 \\
Re\textsuperscript{3}Sum &  36.3 & 24.0 &  36.0 \\
S2S        & 37.6 & 24.6 & 37.3 \\
Uni-model  & 37.9 & 25.0 & 37.6 \\
Concat-S2S & 34.2 & 20.3 & 34.3 \\
Concat-Uni & 37.4 & 24.3 & 36.9 \\
PESG       & \textbf{40.2} & \textbf{28.1} & \textbf{39.9} \\
\midrule
\sbbkgrnd Proto & \sbbkgrnd -- & \sbbkgrnd -- & \sbbkgrnd --  \\
\bottomrule
\end{tabular}
\caption{ROUGE scores comparison with baselines. \texttt{Proto} directly copies from the prototype summary as generated summary.}
\label{tab:comp_rouge_baselines}
\end{table}

\begin{table}[t]
\centering
\small
\begin{tabular}{@{}lcc@{}}
\toprule
& Fluency & Consistency \\
\midrule
Uni-model & 1.61 & 1.53 \\
\cbkgrnd Re\textsuperscript{3}Sum & \cbkgrnd 1.53 & \cbkgrnd 1.14 \\
PESG & \textbf{1.86}\dubbelop & \textbf{1.73}\dubbelop  \\
\bottomrule
\end{tabular}
\caption{Fluency and consistency comparison by human evaluation.}
\label{tab:comp_human_baslines}
\end{table}

We compare our model with the baselines listed in Table~\ref{tab:comp_rouge_baselines}.
Our model performs consistently better than other summarization models including the state-of-the-art model with improvements of 6\%, 12\% and 6\% in terms of ROUGE-1, ROUGE-2 and ROUGE-L.
This demonstrates that prototype document-summary pair provides strong guidance for summary generation that cannot be replaced by other complicated baselines without prototype information.
Meanwhile, directly concatenating the prototype summary with the original input does not increase performance, instead leading to drops of 9\%, 17\%, 8\% and 1\%, 3\%, 2\% in terms of ROUGE 1,2,L on the S2S and Unified models, respectively.
As for the baseline model \texttt{Proto}, we found that it directly \textbf{copies} from the prototype summary as generated summary, which leads to a totally useless and incorrect summary.

For the human evaluation, we asked annotators to rate each summary according to its consistency and fluency.
The rating score ranges from 1 to 3, with 3 being the best.
Table~\ref{tab:comp_human_baslines} lists the average scores of each model, showing that PESG outperforms the other baseline models in both fluency and consistency.
The kappa statistics are 0.33 and 0.29 for fluency and consistency respectively, and that indicates the moderate agreement between annotators.
To prove the significance of these results, we also conduct the paired student t-test between our model and \texttt{Re\textsuperscript{3}Sum} (row with shaded background).
We obtain a p-value of $2 \times 10^{-7}$ and $9 \times 10^{-12}$ for fluency and consistency, respectively.

We also analyze the effectiveness of performance by the two hyper-parameters: $\eta$ and $\epsilon$.
It turns out that our model has a consistently good performance, with ROUGE-1, ROUGE-2, ROUGE-L scores above 39.5, 27.5, 39.4, which demonstrates that our model is very robust.

\subsection{Ablation Study}
\label{sec:ablation}

\begin{figure}[htb] 
  \begin{minipage}[b]{0.45\columnwidth} 
    \centering 
    \includegraphics[width=3.5cm, height=3cm]{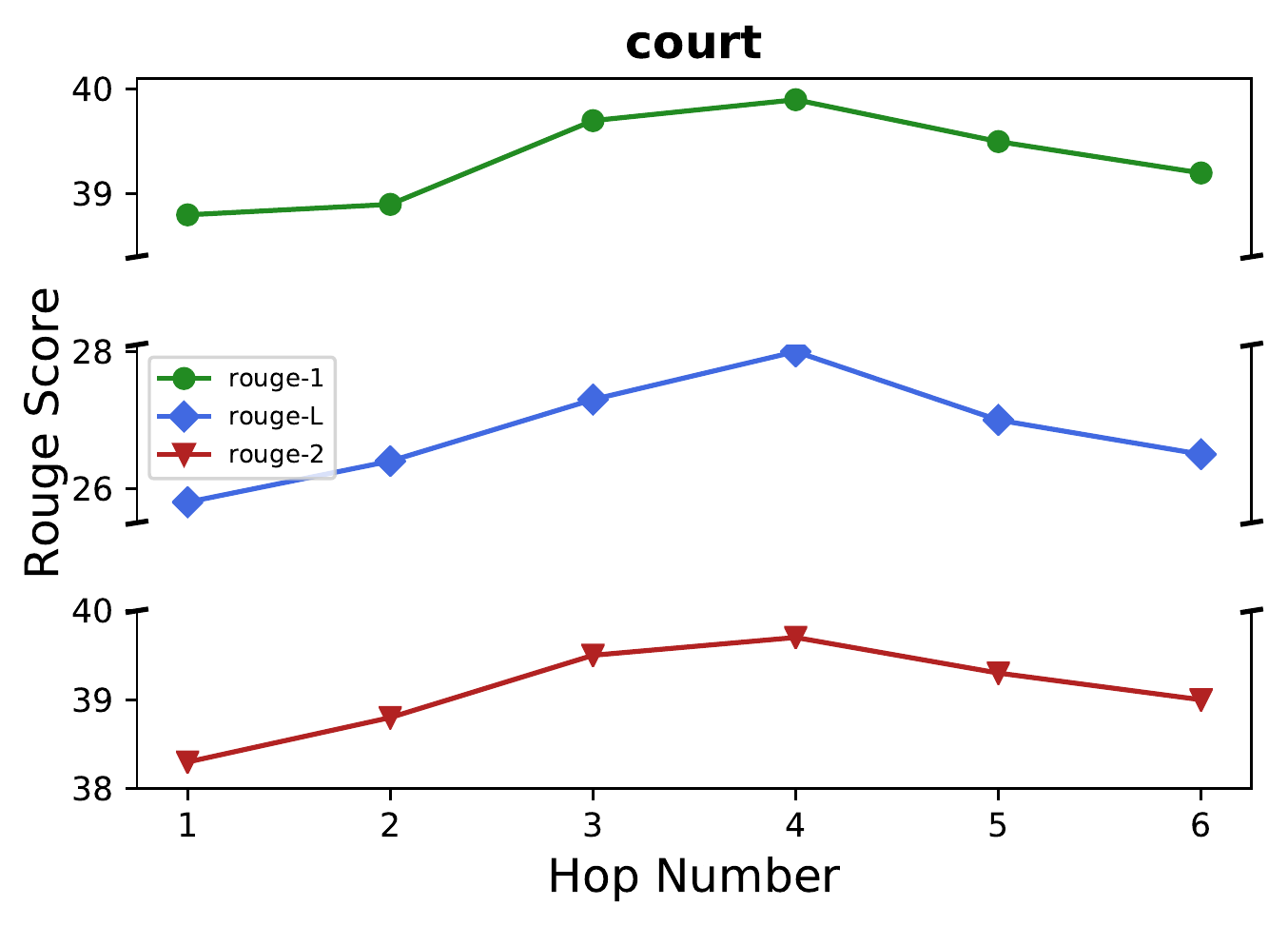}
    \caption{Visualizations of ROUGE score with different hop numbers.}
    \label{fig:court-hop}
  \end{minipage}%
  \quad
  \begin{minipage}[b]{0.45\columnwidth} 
    \centering
    \scriptsize
    \begin{tabular}{@{}lccc@{}}
        \toprule
        & R1 & R2 & RL \\
        \midrule
        PESG-FC & 38.7 & 26.2 & 38.6 \\
        PESG-PR & 37.3 & 24.6 & 37.0   \\
        PESG-SS & 38.6 & 25.6 & 38.3 \\
        PESG-FG & 38.8 & 25.9 & 38.5 \\
        \bottomrule
    \end{tabular}
    \vspace{1cm}
    \tabcaption{ROUGE scores of different ablation models of PESG.\label{tab:comp_bleu_ablation}}
  \end{minipage} 
\end{figure}

The ROUGE scores of different ablation models are shown in Table~\ref{tab:comp_bleu_ablation}.
All ablation models perform worse than PESG in terms of all metrics, which demonstrates the preeminence of PESG. 
More importantly, by this controlled experiment, we can verify the contributions of each modules in PESG.

\subsection{Analysis of Editing Generator}

We visualize the editing gate (illustrated in Equation~\ref{editing-gate}) of two \textbf{randomly sampled} examples, shown in Figure~\ref{fig:edit-gate-vis}.
A lower weight (lighter color) means that the word is more likely to be copied from the summary pattern; that is to say, this word is a universal patternized word.
We can see that the phrase \begin{CJK*}{UTF8}{gkai}本院认为\end{CJK*} (the court held that) has a lower weight than the name of the defendant (PERS), which is consistent with the fact that (the court held that) is a patternized word and the name of the defendant is closely related to the input document.

We also show a case study in Table~\ref{tab:case}, which includes the input document and reference summary with the generated summaries.
Underlined text denotes a grammar error and a strike-through line denotes a fact contrary to the input document.
We only show part of the document and summary due to limited space; the full version is shown in Appendix.
As can be seen, the summary generated by \texttt{Uni-model} faces an inconsistency problem and the summary generated by \texttt{Re\textsuperscript{3}Sum} is contrary to the facts described in the input document.
However, PESG overcomes both of these problems and generates an accurate summary with good grammar and logic.

\subsection{Analysis of Fact Extraction Module}
\label{sec:fgru-hop}

We investigate the influence of the iteration number when facts are extracted.
Figure~\ref{fig:court-hop} illustrates the relationship between iteration number and the f-value of the ROUGE score.
The results show that the ROUGE scores first increases with the number of hops. 
After reaching an upper limit it then begins to drop.
This phenomenon demonstrates that the fact extraction module is effective by polishing the facts representation.

\begin{CJK*}{UTF8}{gkai}
\begin{table}[t]
\centering
\scriptsize
\begin{tabular}{l|l}
\toprule
\multicolumn{2}{p{7.6cm}}{安徽省 人民法院 作出 刑事 判决 ， 以 斗殴 罪 判处 被告人 PERS 有期徒刑 YEARS 。  合肥市 监狱 提出 减刑 意见 。  执行 机关 以 PERS 在 服刑 期间 确 有 悔改 表现 ， 已 获得 奖励 二 个 ， 建议 对 其 减刑 。 经 审理 查明 ， 罪犯 PERS 在 服刑 期间 ， 能 认罪 服法 , 努力 参加 技术 学习 。(Anhui Provincial People's Court made a criminal judgment and sentenced the defendant PERS to YEARS imprisonment for the crime of fighting. Hefei Prison proposed a commutation sentence since PERS had repentance and received two awards during his sentence. It was found through trial that the criminal PERS pleaded guilty and strived to participate in technical learning while serving his sentence.)}     \\ 
\hline
\rotatebox{270}{reference}                       & \multicolumn{1}{p{7cm}}{本院 认为， 罪犯 PERS 在 服刑 期间 确 有 悔改，表现符合 法定 减刑 条件。依据 《 中华人民共和国 刑法 》 第 NUM 款 之 规定，对 罪犯 PERS 减去 自 本 裁定 送达 之 日 起 至 DATE 止 尚未 执行 完毕 的 刑罚。本 裁定 送达 后 即 发生 法律 效力。 (The court believed that the criminal PERS did have repentance during his sentence, and his performance was in line with the statutory commutation conditions. According to the provisions of Section NUM of the Criminal Law of the People's Republic of China, the court decided to exempt PERS from penalty that had not been executed since the date of delivery of this ruling until DATE.
Legal effect would occur upon the delivery of this ruling.)}   \\ \hline
\rotatebox{270}{\texttt{Uni-model}}           & \multicolumn{1}{p{7cm}}{本院 认为 ， 被告人 PERS \sout{以 非法 占有 为 目的 ， 秘密 窃取 他人 财物} ， \sout{数额 较 大 ， 其 行为 已 构成 盗窃罪}。 公诉 机关 指控 被告人 PERS勋 斗殴 罪 的 事实 清楚 ， 证据 确实 充分 ， 本院 予以 支持。 (The court held that the defendant PERS \sout{secretly stole the property of others for the purpose of illegal possession, the stolen amount was relatively large, and his behavior had constituted theft.} The fact that defendant PERS committed the crime of fighting was clear, the evidence was indeed sufficient, and the court supported it.)}    \\ \hline
\rotatebox{270}{\texttt{Re\textsuperscript{3}Sum}} & \multicolumn{1}{p{7cm}}{本院 认为 , PERS \sout{在 道路 上 驾驶 机动车},其 行为 已 构成 \uline{上诉人 服刑} \uline{确 系 罪}. 公诉 机关 指控 的 犯罪 事实 清楚，证据 确实 充分。 (The court held that PERS was \sout{driving a motor vehicle on the road}, and his behavior constituted that \ul{the appellant was guilty of serving a sentence}. The criminal facts accused by the public prosecution agency were clear and the evidence was indeed sufficient.)}  \\ \hline
\rotatebox{270}{PESG}                      & \multicolumn{1}{p{7cm}}{本院 认为，罪犯 PERS 在 服刑 期间，确 有 悔改 表现。符合 法定 减刑 条件，依照 《 中华人民共和国 刑法 》 第 NUM 条 、 第 NUM 条 之 规定，裁定 如下 对 罪犯 PERS 减去 有期徒刑 MONTHS。本 裁定 送达 后 即 发生 法律 效力。 (The court held that the criminal PERS did have repentance during his sentence. In accordance with the provisions of NUM and NUM of the Criminal Law of the People's Republic of China, the ruling was as follows: exempted PERS from the MONTHS penalty. Legal effect would occur upon the delivery of this ruling.)} \\ 
\bottomrule
\end{tabular}
\caption{Examples of the generated natural answers by PESG and other models.}
\label{tab:case}
\end{table}
\end{CJK*}

\section{Conclusion}

In this paper, we propose a framework named \emph{Prototype Editing based Summary Generator} (PESG), which aims to generate summaries in formal writing scenarios, where summaries should conform to a patternized style.
Given a prototype document-summary pair, our model first calculates the cross dependency between the prototype document-summary pair.
Next, a fact extraction module is employed to extract facts from the document, which are then polished.
Finally, we design an editing-based generator to produce a summary by incorporating the polished fact and summary pattern.
To ensure that the generated summary is consistent with the input document, we propose a fact checker to estimate the mutual information between the input document and generated summary.
Our model outperforms state-of-the-art methods in terms of ROUGE scores and human evaluations by a large margin, which demonstrates the effectiveness of PESG. 

\section*{Acknowledgments}
We would like to thank the anonymous reviewers for their constructive comments. 
We would also like to thank Anna Hennig in Inception Institute of Artificial Intelligence for her help on this paper. 
This work was supported by the National Key Research and Development Program of China (No. 2017YFC0804001), the National Science Foundation of China (NSFC No. 61876196 and NSFC No. 61672058)

\appendix
\section{SRU Cell}
\label{sec:fgru}

Gated recurrent unit (GRU)~\cite{Cho2014LearningPR} is a gating mechanism in recurrent neural networks, which incorporate an update gate in an RNN.
We first give the details of the original GRU here.
\begin{align}
u_{i} &= \sigma(W^{(u)}x_{i}+U^{(u)}h_{i-1}+b^{(u)}),  \label{gru-gated} \\
r_{i} &= \sigma(W^{(r)}x_{i}+U^{(r)}h_{i-1}+b^{(r)}),  \\
\tilde{h_{i}} &= \text{tanh}(W^{(h)}x_{i}+r_{i}\circ Uh_{i-1}+b^{(h)}), \\
h_{i} &= u_{i}\circ\tilde{h_{i}}+(1-u_{i})\circ h_{i-1}, \label{gru-hi}
\end{align}
where $\sigma$ is the sigmoid activation function, $W^{(u)}, W^{(r)}, W^{(h)} \in \mathbb{R}^{n_{H}\times n_{I}},U^{(u)},U^{(r)},U\in \mathbb{R}^{n_{H}\times n_{H}}, n_{H}$ is the hidden size, and $n_{I}$ is the size of input $x_{i}$.
In the original version of GRU, the update gate $u_i$ in Equation~\ref{gru-gated} is used to decide how much of the hidden state should be retained and how much should be updated. 
In our case, we want to decide which facts are salient according to the polished facts $q_{k-1}$ at the $k$-th hop.
To achieve this, we replace the calculation of $u_i$ with the newly computed update gate $g_i$:
\begin{align}
f_{i} &= [x_{i} \circ q_{k-1}; x_{i}; q_{k-1}], \\
z_{i} &= W^{(2)}\tanh(W^{(1)}f_{i}+b^{(1)})+b^{(2)}, \\
g_{i} &= \frac{\exp(z_{i})}{\sum^{n_{s}}_{j=1} \exp(z_{j})},
\end{align}
where $W^{(2)}, W^{(1)}, b^{(1)}, b^{(2)}$ are all trainable parameters and $k$ is the hop number in the multi-hop situation which is a hyper-parameter manually set.
The effectiveness of this hyper-parameter is verified in the experimental results shown in \S~\ref{sec:fgru-hop}.
Equation~\ref{gru-hi} now becomes:
\begin{align}
h_{i} &= g_{i} \circ \tilde{{h_{i}}}+(1-g_{i}) \circ h_{i-1 }.
\end{align}
We use the name ``SRU'' to denote this modified version of an GRU cell.

\section{Case Study}
\vspace{-1cm}
\label{sec:appendix_case}

\begin{CJK*}{UTF8}{gkai}
\begin{table}[b]
\centering
\vspace{-0.3cm}
\scriptsize
\begin{tabular}{l|l}
\toprule
\multicolumn{2}{p{7.4cm}}{安徽省 合肥市 包河区 人民法院 作出 刑事 判决 ， 以 斗殴 罪 判处 被告人 PERS 有期徒刑 YEARS 。 案经 本院 二审 ， 准许 上诉人 撤回 上诉 。 判决 生效 后 交付 执行 。 执行 机关 合肥市 义城 监狱 于 DATE 提出 减刑 意见 ， 报送 本院 审理 。 本院 依法 组成 合议庭 进行 了 审理 ， 现 已 审理 终结 。 执行 机关 以 罪犯 PERS 在 服刑 期间 确 有 悔改 表现 ， 已 获得 奖励 二 个 ， 建议 对 其 减刑 。 经 审理 查明 ， 罪犯 PERS 在 服刑 期间 ， 能 认罪 服法 ， 努力 参加 技术 学习 ， 积极 完成 生产 任务 。 (The People's Court of Baohe District, Hefei City, Anhui Province made a criminal judgment and sentenced the defendant PERS to YEARS imprisonment for the crime of fighting. In the second instance of the court, the appellant was allowed to withdraw the appeal. The judgment takes effect after being delivered. The enforcement agency Hefei City Yicheng Prison put forward a commutation opinion on DATE and submitted it to the court for trial. The court had formed a collegiate bench in accordance with the law and had now end the trial. Criminal PERS had repentance during his sentence and received two awards, thus the enforcement agency suggested he should be commuted. Investigation confirmed that the criminal PERS was able to plead guilty while serving his sentence, strived to participate in technical studies, and completed production tasks.)}     \\ 
\hline
\rotatebox{270}{reference}                                     & \multicolumn{1}{p{6.6cm}}{本院 认为， 罪犯 PERS 在 服刑 期间 确 有 悔改 表现符合 法定 减刑 条件。依据 《 中华人民共和国 刑法 》 第 NUM 款 之 规定，裁定 如下对 罪犯 PERS 减去 自 本 裁定 送达 之 日 起 至 DATE 止 尚未 执行 完毕 的 刑罚。本 裁定 送达 后 即 发生 法律 效力。 (The court held that the criminal PERS did have repentance during the period of serving his sentence in accordance with the statutory commutation conditions. According to the provisions of NUM Section of the Criminal Law of the People's Republic of China, the court decided to exempt PERS from penalty that had not been executed since the date of delivery of this ruling until DATE. Legal effect would occur upon the delivery of this ruling.)}   \\ \hline
\rotatebox{270}{Unified}                                          & \multicolumn{1}{p{6.6cm}}{本院 认为 ， 被告人 PERS 以 非法 占有 为 目的 ， 秘密 窃取 他人 财物 ， 数额 较 大 ， 其 行为 已 构成 盗窃罪 ， 依法 应 予 惩处。 公诉 机关 指控 被告人 PERS 聚众斗殴 罪 的 事实 清楚 ， 证据 确实 充分 ， 指控 罪名 成立 ， 本院 予以 支持。 被告人 PERS 曾 因 故意 犯罪 被 判处 有期徒刑 ， 在 刑罚 执行 完毕 后 YEARS 内 再 犯， 应当 判处 有期徒刑 以上 刑罚 之 罪 ， 是 累犯 ， 依法 应当 从重 处罚 。 (The court held that the defendant PERS secretly stole the property of others for the purpose of illegal possession, and the amount was relatively large. His behavior had constituted theft and should be punished according to law. The fact that the defendant PERS commiting the crimes of the fighting was clear, the evidence was indeed sufficient, thus the charges were established and the court supported it. The defendant PERS was sentenced to fixed-term imprisonment for intentional crimes. If he committed another crime in YEARS after the execution of the penalty, he should be sentenced to a fixed-term imprisonment or more. He is a recidivist and should be severely punished according to law.)}    \\ \hline
\rotatebox{270}{Re\textsuperscript{3}Sum}                                          & \multicolumn{1}{p{6.6cm}}{本院 认为NUM PERS 在 道路 上 驾驶 机动车其 行为 已 构成 上诉人 服刑 确 系 罪公诉 机关 指控 的 犯罪 事实 清楚证据 确实 充分指控 的 罪名 成立本院 予以 支持被告人 PERS 归案 后 如实 供述 自己 的 罪行依法 予以 从轻 处罚公诉 机关 的 量刑 建议 适当本院 予以 采纳 此依照 《 中华人民共和国 刑法 》 第 NUM 款 之 规定判决 如下被告人 PERS 犯 服刑判处 有期徒刑 MONTHS并 处 罚金 人民币 MONEY (The court held that NUM PERS was driving a motor vehicle on the road, and his behavior constituted that the appellant was guilty of serving a sentence. The criminal facts accused by the public prosecution agency were clear, the evidence was indeed sufficient, and the charges were supported by the court. After confessing his crimes, he should be given a lighter punishment according to law. The public prosecution agency's sentencing recommendations were appropriate and adopted by the court. According to the provisions of Section NUM of the Criminal Law of the People's Republic of China, the judgment was as follows, the defendant PERS was sentenced to imprisonment and sentenced to fixed-term MONTHS imprisonment, and fined the penalty RMB MONEY.)}  \\ \hline
\rotatebox{270}{PESG}                                          & \multicolumn{1}{p{6.6cm}}{本院 认为，罪犯 PERS 在 服刑 期间，确 有 悔改 表现。符合 法定 减刑 条件，依照 《 中华人民共和国 刑法 》 第 NUM 条 、 第 NUM 条 之 规定，裁定 如下：对 罪犯 PERS 减去 有期徒刑 MONTHS。本 裁定 送达 后 即 发生 法律 效力。 (The court believed that the criminal PERS did have repentance during his sentence. In accordance with the statutory commutation conditions and NUM and NUM of the Criminal Law of the People's Republic of China, the ruling was as follows: exempted PERS from the MONTHS penalty. Legal effect would occur upon the delivery of this ruling.)} \\ 
\bottomrule
\end{tabular}
\caption{Examples of the generated natural answers by PESG and other models.}
\label{tab:appendix_case}
\end{table}
\end{CJK*} 
\clearpage
\bibliographystyle{acl_natbib}
\bibliography{references}

\end{document}